\documentclass[runningheads]{llncs}
\usepackage{graphicx}
\usepackage{amsmath,amssymb} 
\usepackage{color}
\usepackage[width=122mm,left=12mm,paperwidth=146mm,height=193mm,top=12mm,paperheight=217mm]{geometry}
\usepackage{subfigure}
\usepackage{multirow}
\usepackage{diagbox}
\usepackage{mathrsfs}
\usepackage{pdfpages}
\usepackage{enumerate}
\usepackage{float}
\newcolumntype{C}[1]{>{\centering}p{#1}}
\newcolumntype{L}[1]{>{\raggedleft}p{#1}}
\newcolumntype{R}[1]{>{\raggedright}p{#1}}
\begin{document}
\mainmatter
\def\ECCV18SubNumber{2287}  

\title{Deep Sampling Networks} 

\titlerunning{Deep Sampling Networks}

\authorrunning{B. Cai et al.}

\author{Bolun Cai$^1$\thanks{The early work was completed while the author was with Tencent Wechat AI.} 
\and Xiangmin Xu$^{1}$\thanks{X. Xu is the corresponding author.}
\and  Kailing Guo$^1$
\and Kui Jia$^1$
\and Dacheng Tao$^2$
}
\institute{$^1$South China University of Technology, China\\
$^2$UBTECH Sydney AI Centre, FEIT, The University of Sydney, Australia\\
\email{caibolun@gmail.com}, \email{\{xmxu,kuijia,guokl\}@scut.edu.cn, \email{dacheng.tao@sydney.edu}}
}

\maketitle

\begin{abstract}
Deep convolutional neural networks achieve excellent image up-sampling performance. However, CNN-based methods tend to restore high-resolution results highly depending on traditional interpolations (e.g. bicubic). In this paper, we present a deep sampling network (DSN) for down-sampling and up-sampling without any cheap interpolation. First, the down-sampling subnetwork is trained without supervision, thereby preserving more information and producing better visual effects in the low-resolution image. Second, the up-sampling subnetwork learns a sub-pixel residual with dense connections to accelerate convergence and improve performance. DSN's down-sampling subnetwork can be used to generate photo-realistic low-resolution images and replace traditional down-sampling method in image processing. With the powerful down-sampling process, the co-training DSN set a new state-of-the-art performance for image super-resolution. Moreover, DSN is compatible with existing image codecs to improve image compression.
\end{abstract}

\section{Introduction}
The aim of the image sampling is to generate a low-resolution (LR) image from a high-resolution (HR) image or reconstruct the HR image in reverse. Single-image up-sampling is widely used in computer vision applications including HDTV \cite{hdtv}, medical imaging \cite{medical}, satellite imaging \cite{satellite}, and surveillance \cite{surveillance}, where high-frequency details are required on demand. As the use of mobile social networks (e.g., Google+, WeChat, and Twitter) continues to grow, thumbnail down-sampling is another important way to optimize data storage and transmission over limited-capacity channels.

Image up-sampling, also known as super-resolution (SR), has been studied for decades. Early methods including bicubic interpolation \cite{bicubic}, Lanczos resampling \cite{lanczos}, gradient profiles \cite{gradient}, and patch redundancy \cite{patch} are based on statistical image priors or internal patch representations. More recently, learning-based methods have been proposed to model a mapping from LR to HR patches such as neighbor embedding \cite{ne}, sparse coding \cite{sc}, and random forests \cite{rf}. However, image up-sampling is highly ill-posed, since the HR to LR process contains non-invertible down-sampling.

Due to their powerful learning capability, deep convolutional neural networks (CNNs) have achieved state-of-the-art performance in many computer vision tasks, such as image classification \cite{imagenet}, object detection \cite{detection}, and image segmentation \cite{deeplab}. Recently, CNNs have been used to address this ill-posed inverse problem, demonstrating superiority over traditional learning paradigms. The first example of the approach, super-resolution convolutional neural network (SRCNN) \cite{srcnn} predicted the nonlinear LR to HR mapping in an end-to-end manner. To reduce computational complexity, fast SRCNN (FSRCNN) \cite{fsrcnn} and efficient sub-pixel convolutional neural network (ESPCN) \cite{espcn} up-scaled the resolution only at the output layer. Kim et al. \cite{vdsr} developed a very deep super-resolution (VDSR) network with 20 convolutional layers by residual learning, and Mao et al. \cite{red} proposed a 30-layer residual encoder-decoder (RED) network with symmetric skip connections to facilitate training. Deeply-recursive convolutional network (DRCN) \cite{drcn} introduced a very deep recursive layer via a chain structure with 16 recursions, and deep recursive residual network (DRRN) \cite{drrn} adopted recursive residual units to control the model parameters while increasing depth.

\begin{figure}[bth]
\begin{center}
   \includegraphics[width=0.8\linewidth]{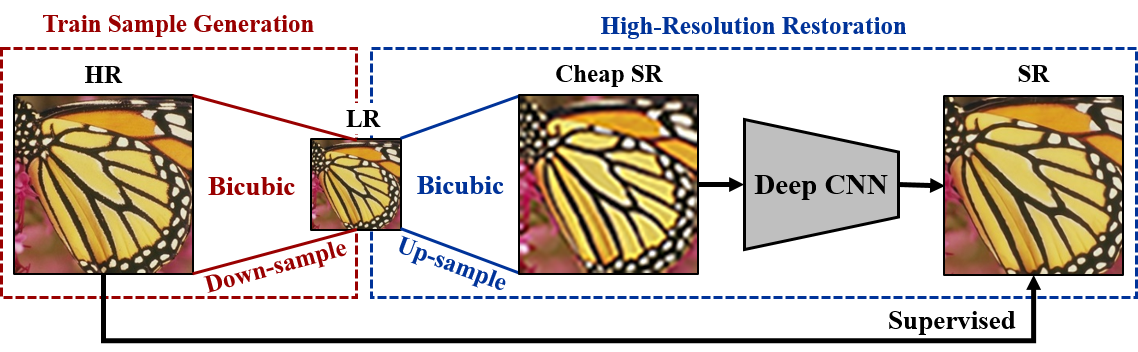}
\end{center}
   \caption{The classical framework based on CNNs for image up-sampling \cite{srcnn,vdsr,red,drcn,drrn}. The HR image is down-sampled to synthesize the training samples. Then, the deep model learns the mapping between the HR and the cheap SR up-scaled by the same factor via bicubic interpolation.}
\label{fig:sr}
\end{figure}

Despite achieving excellent performance, the CNN-based methods are highly dependent on interpolation-based down/up-sampling, as shown in Fig. \ref{fig:sr}. The limitations of these methods arise from two aspects:\\
\\
\textbf{Down-sampling.} For down-sampling, the model trained for a specific interpolation does not work well with the other interpolations, and different interpolations significantly alter restoration accuracy. Therefore, all the solutions mentioned above generally assume that the degradation is a bicubic interpolation (the default setting of $imresize()$ in \emph{Matlab}) when shrinking an image. However, the bicubic function based on a weighting transformation resprents a cheap down-sampling process that discards useful high-frequency details for HR image restoration.\\
\textbf{Up-sampling.} With cheap up-sampling, the networks \cite{vdsr,red,drcn,drrn} increase the resolution during preprocessing or at the first layer to learn the interpolation's residual. However, the up-sampling do not add information to solve the ill-posed restoration problem. To replace cheap interpolation, ESPCN \cite{espcn} and FSRCNN \cite{fsrcnn} adopt sub-pixel shuffling and a deconvolution layer to improve efficiency, respectively. However, without cheap up-sampling, they carry the input and restore the details as an auto-encoder, so converge slowly.

To address the above problems, we propose a deep sampling network (DSN) without any cheap interpolation that is trained for simultaneous down- and up-sampling.
\begin{itemize}
  \item A learnable down-sampling subnetwork (Down-SNet) is trained without supervision, thereby preserving more information and transmitting a better visual effect to the LR image. For self-supervision, the super-pixel residual is adopted with a novel activation function, called quantized bilateral ReLU (Q-BReLU).
  \item An up-sampling subnetwork (Up-SNet) learns a sub-pixel residual with dense connections to accelerate convergence and improve performance. First, the dense pixel representation trained with deep supervision extracts the multi-scale features by multi-level dictionaries, and second, the sub-pixel residual restores the HR result without cheap up-sampling.
\end{itemize}
Compare to traditional down-sampling methods, the Down-SNet can preserve more useful information and generate photo-realistic LR images. Compared to existing CNN-based up-sampling methods, the co-training DSN achieves the best performance with lower computational complexity. 

\section{CNN-based Sampling in Related Works}
We first design an experiment to investigate the sampling in CNN-based SR methods. In this section, we re-implement\footnote{The SRCNN (9-1-5) re-implemented here achieves better perform than reported by the authors (32.39dB) \cite{srcnn}.} the baseline SRCNN \cite{srcnn} model with the Adam \cite{adam} optimizer. The learning rate decreases by a factor of 0.1 from $10^{-3}$ to $10^{-5}$ every 50 epochs.
\begin{figure}[bth]
\begin{center}
\subfigure[Down-sampling]{\label{subfig:downsample}\includegraphics[width=0.4\linewidth]{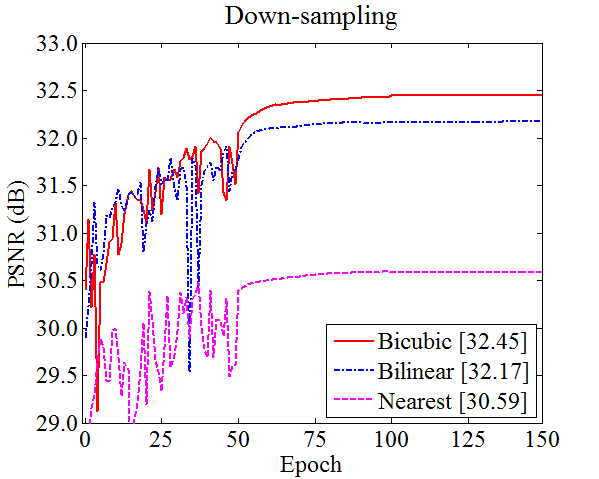}}
\subfigure[Up-sampling]{\label{subfig:upsample}\includegraphics[width=0.4\linewidth]{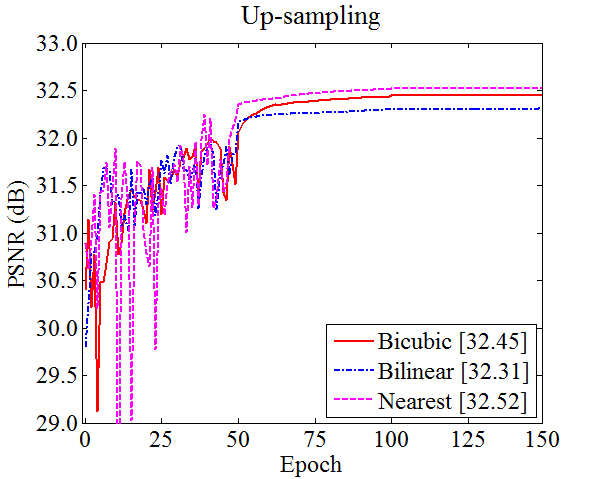}}
\end{center}
   \caption{Convergence and accuracy analyses on different down/up-sampling methods.}
\label{fig:sample}
\end{figure}

\subsection{Learn Restoration for Specified Down-sample}
Almost all CNN-based SR methods \cite{srcnn,fsrcnn,espcn,red,vdsr,drcn,drrn,memnet} are trained for a specified down-sampling. As shown in Table \ref{tab:sp}, down-sampling asymmetry in the training/testing phase results in poor restorations, even worse than the direct bicubic interpolation. This is because the CNNs learn the targeted mapping for the specified degradation. Moreover, different down-sampling (nearest-neighbor, bilinear, and bicubic) produce significantly different convergence rates and restoration accuracies shown in Fig. \ref{subfig:downsample}. The degradation with bicubic down-sampling contains more useful information, so the bicubic model converges faster and achieves better performance. However, the bicubic interpolation is still a cheap down-sampling process that discards useful details for HR image restoration. In this paper, we simultaneously train a deep sampling network for down-sampling and corresponding up-sampling.
\begin{table}[th]
\small
\center
\caption{Comparison of different down-sampling degradations for scale factor $\times 3$ on Set5 \cite{set5} with PSNR (dB). The baseline PSNR of bicubic interpolation is 30.39.}
\renewcommand\arraystretch{1.1}
\begin{tabular}{|c|C{1.5cm}C{1.5cm}C{1.5cm}|c|}
\hline
\diagbox{Train}{Test} & Nearest & Bilinear & Bicubic & ~~~~Avg.~~~~\\\hline
Nearest  & \textbf{30.59} & 29.76 & 30.56 & 30.30 \\\hline
Bilinear & 25.38 & \textbf{32.17} & 31.19 & 29.58 \\\hline
Bicubic  & 27.59 & 31.92 & \textbf{32.45}\protect\footnotemark[1] & \textbf{30.70} \\\hline
\end{tabular}\label{tab:sp}
\end{table}

\subsection{Learn Mapping from Cheap Up-sample}

In the popular CNN-based SR methods (e.g. SRCNN \cite{srcnn}, VDSR \cite{vdsr}, DRCN \cite{drcn}, and DRRN \cite{drrn}), cheap up-sampling is used during preprocessing to increase the resolution before or at the first network layer. As shown in Fig. \ref{subfig:upsample}, interpolations do not add information to improve restoration accuracy. Instead of improving accuracy, the complex (bicubic) preprocessing prematurely introduces smooth and inaccurate interpolation, resulting in a hard-to-train network. Conversely, the nearest-neighbor interpolation achieves the best performance because it selects the raw value of the nearest point and does not consider the values of neighboring points. To address this problem, ESPCN \cite{espcn} and FSRCNN \cite{fsrcnn} carry the input and restore details without cheap up-sampling. However, deep information carry results in a decreasing convergence rate. Therefore, we propose sub-pixel residual learning to accelerate convergence and improve performance.

\section{Deep Sampling Network}
\begin{figure*}[th]
\begin{center}
   \includegraphics[width=1\linewidth]{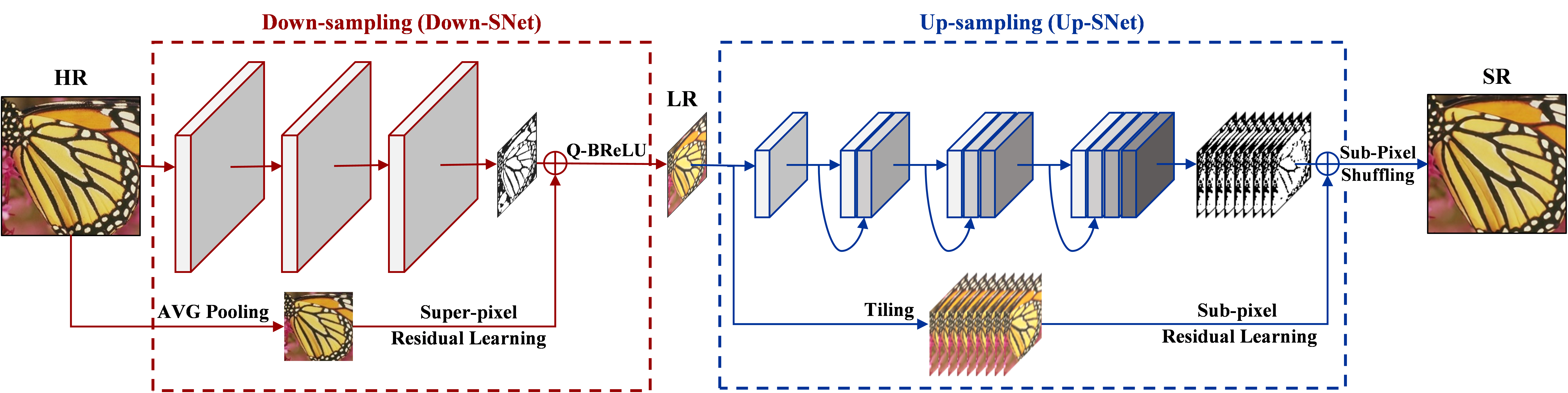}
\end{center}
   \caption{The deep sampling network. (1) The down-sampling subnetwork (Down-SNet) is trained in an unsupervised manner with super-pixel residual learning and the Q-BReLU function. (2) The up-sampling subnetwork (Up-SNet) learns the dense pixel representation by sub-pixel residual learning.}
\label{fig:rrsr}
\end{figure*}

In this paper, we propose a deep sampling network (DSN) composed of a down-sampling subnetwork and an up-sampling subnetwork. The DSN architecture is presented in Fig. \ref{fig:rrsr}.

\subsection{Unsupervised Down-sampling Subnetwork}

A learnable down-sampling subnetwork (Down-SNet) is trained without supervision, which learns the super-pixel residual with a novel activation function, called quantized bilateral ReLU (Q-BReLU).

\subsubsection{Super-pixel Residual Learning}

Without a supervised signal, Down-SNet can proactively retain useful information and discard redundant information. However, the multi-layer network is an end-to-end relationship requiring very long-term memory. For this reason, the LR image generated from the learned features contains artifacts. We can simply solve this problem by super-pixel residual-learning.

In Down-SNet, the pixel of LR output $\textbf{L}_{x,y}$ of scale factor $1/s$ is largely similar to the super-pixel of the HR input $\textbf{H}$. Therefore, we define the super-pixel residual image for down-sampling $\textbf{R}_{x,y}^d = {\textbf{L}_{x,y}} - \frac{1}{{\left| \Omega  \right|}}\sum\nolimits_{\left\{ {m,n} \right\} \in {\Omega _{s \cdot \left( {x,y} \right) - \left\lfloor {s/2} \right\rfloor }}} {{\textbf{H}_{m,n}}}$, where $\left\lfloor{\cdot}\right\rfloor$ reduces the integer and $\Omega$ is a neighborhood with the size of $s\times s$. In $\textbf{R}^d$, most values are likely to be very low and even close to zero. Formally, the down-sampling is denoted $\mathcal{F}^d\left[\textbf{H}\right] \rightarrow \textbf{R}^d$, which includes an inference model (three convolutional layers of size $3\times 3$ and stride $1$) and a down-sampling layer (a convolutional layer of size $s\times s$ and stride of $s$). The original mapping is recast into
 \begin{equation}\label{L}
 {\textbf{L}_{x,y}} = {\mathcal{F}^d}{\left[ \textbf{H} \right]_{x,y}} + \frac{1}{{\left| \Omega  \right|}}\sum\nolimits_{\left\{ {m,n} \right\} \in {\Omega _{s \cdot \left( {x,y} \right) - \left\lfloor {s/2} \right\rfloor }}} {{\textbf{H}_{m,n}}},
 \end{equation}
which can be implemented by feedforward neural networks with shortcut connections \cite{resnet} and average pooling.

\subsubsection{Quantized Bilateral ReLU (Q-BReLU)}
Standard nonlinear activation functions such as the rectified linear unit (ReLU) offer local linearity to overcome the vanishing gradient problem. However, ReLU is designed for classification problems rather than image restoration. In particular, ReLU only inhibits values less than zero, which might lead to response overflow especially without supervision. Moveover, the general digital image is quantified to integers between 0 and 255.

\begin{figure}[th]
\begin{center}
  \subfigure[Q-BReLU]{\label{subfig:qbrelu}\includegraphics[width=0.4\linewidth]{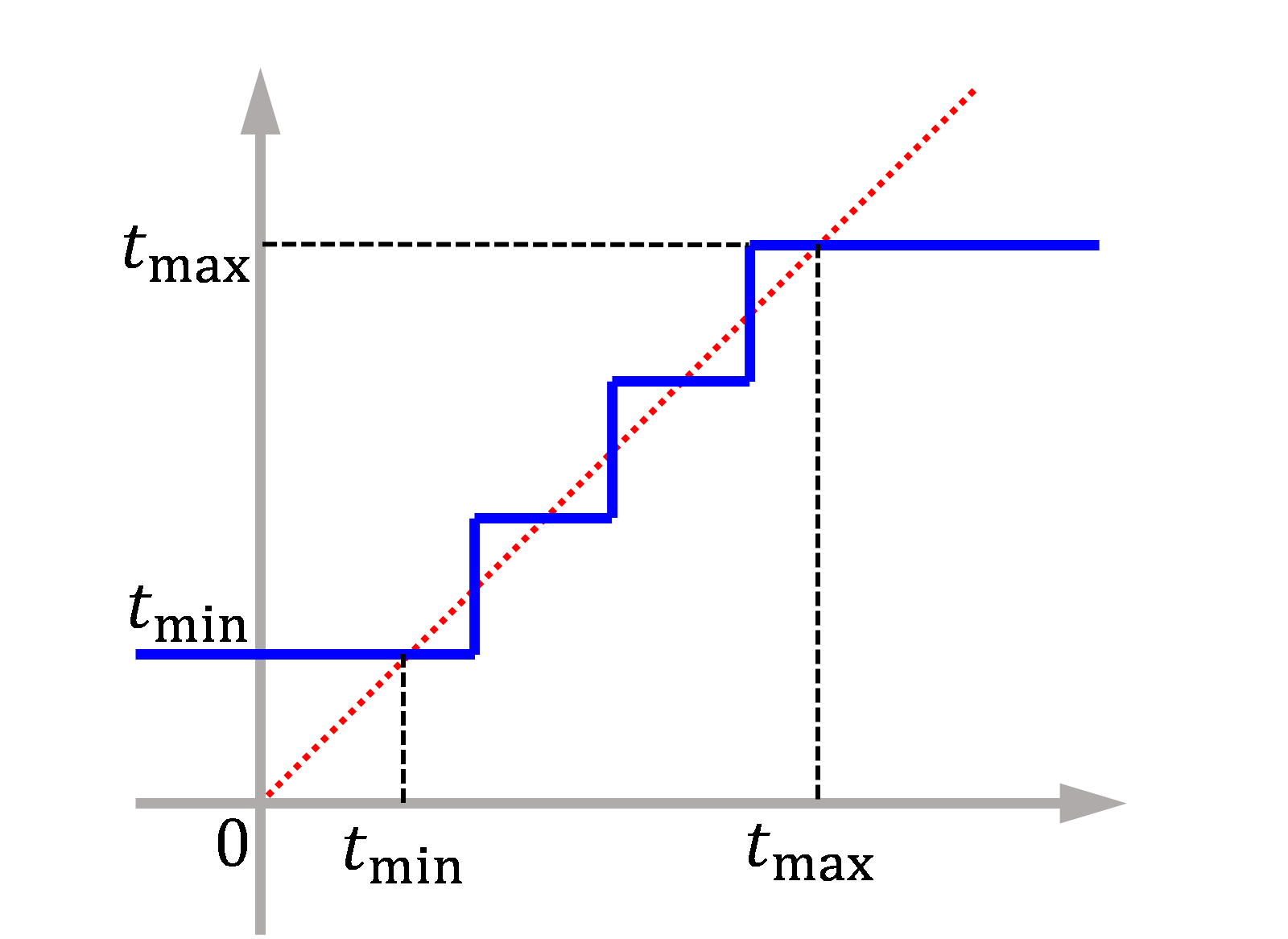}}
  \subfigure[The gradient of Q-BReLU]{\label{subfig:gqbrelu}\includegraphics[width=0.4\linewidth]{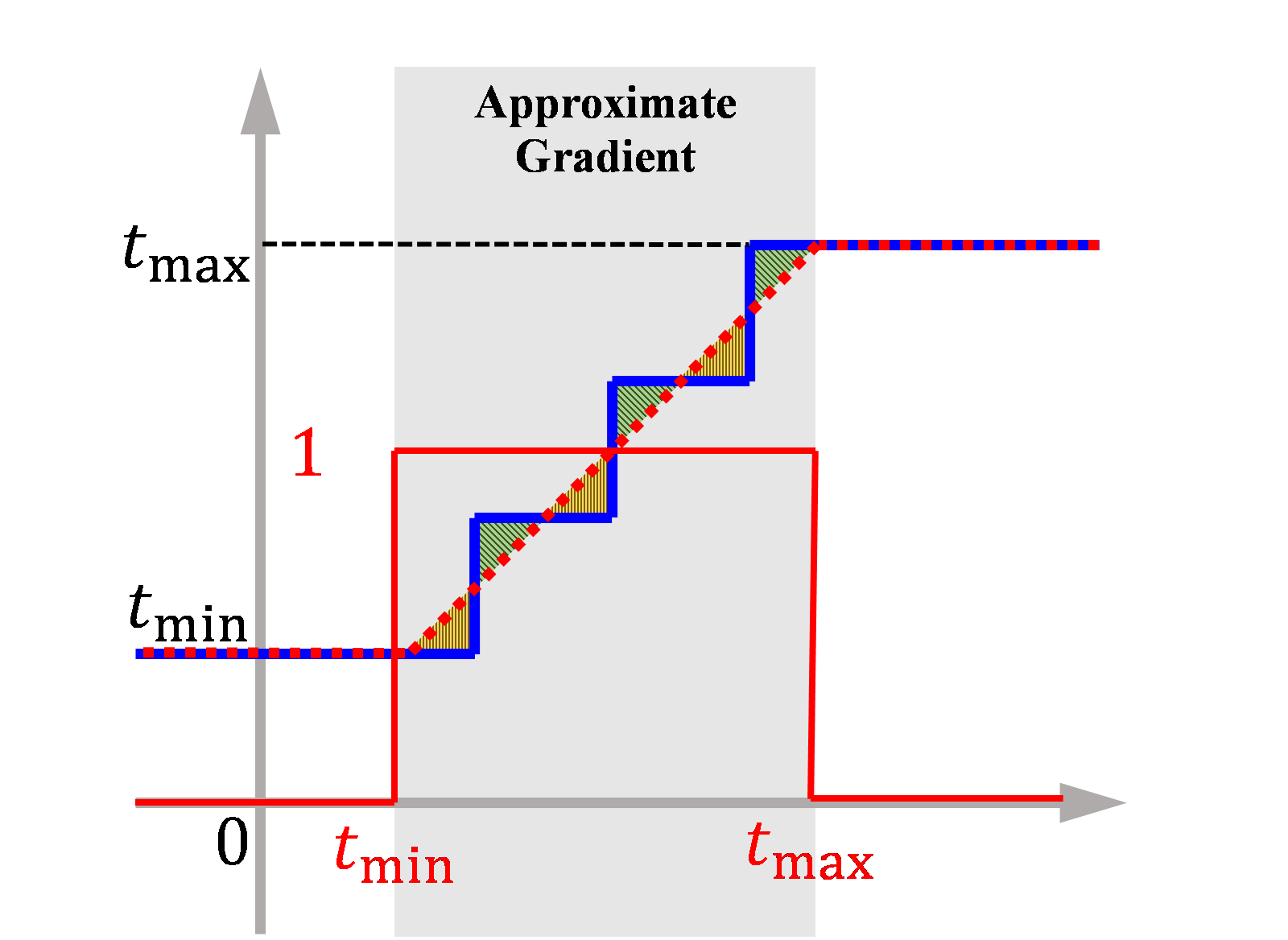}}
\end{center}
   \caption{The quantized bilateral rectified linear unit (Q-BReLU) with 2-bit quantities $Q=2^{2}$. (a) Q-BReLU is denoted in \emph{\textbf{\color{blue}{solid blue}}}. (b) The approximate gradient of Q-BReLU is denoted in \emph{\textbf{\color{red}{dashed red}}}, and \emph{\textbf{\color{red}{solid red}}} denotes the spline fitting Q-BReLU.\protect\\ \textbf{Center quantization}: the high-precision value is rounded to the nearest quantization interval (between neighboring blue dashes); \textbf{Zero mean deviation}: the positive/negative (green/yellow area) deviation balance out the approximate bias.}
\label{fig:qbrelu}
\end{figure}

To overcome this limitation, here we propose the quantized bilateral rectified linear unit (Q-BReLU) to keep bilateral restraint and response quantization, as shown in Fig. \ref{fig:qbrelu}. Q-BReLU is a variation of BReLU \cite{dehazenet}, which is adopted for haze transmission restoration. BReLU is defined as ${f_{brelu}} = \max \left( {\min \left( {x,{t_{\max }}} \right),{t_{\min }}} \right)$, where $t_{\min,\max}$ is the marginal value. Denoting $\Delta t=t_{\max}-t_{\min}$ for terse expression, Q-BReLU is defined as
\begin{equation}\label{qbrelu}
\begin{array}{l}
{f_{qbrelu}}\left( x \right) = \\
\quad \;\;\;\frac{{\Delta t}}{{Q - 1}}\left\lfloor {\frac{{Q - 1}}{{\Delta t}}\left( {{f_{brelu}}\left( x \right) - {t_{\min }}} \right) + 0.5} \right\rfloor  + {t_{\min }},
\end{array}
\end{equation}
where $Q$ is the number of quantities.

\begin{figure}[th]
\center
\subfigure[HR]{\includegraphics[width=0.21\linewidth]{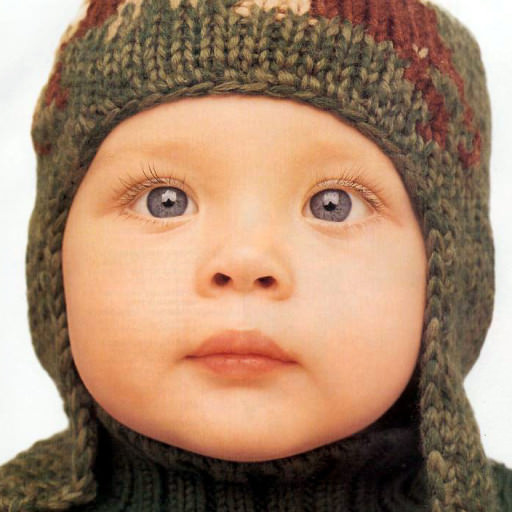}}
\subfigure[Bicubic]{\includegraphics[width=0.21\linewidth]{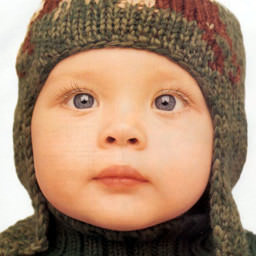}}
\subfigure[w/o Q-BReLU]{\includegraphics[width=0.21\linewidth]{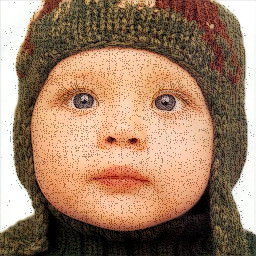}}
\subfigure[Q-BReLU]{\includegraphics[width=0.21\linewidth]{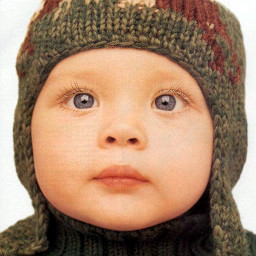}}
\caption{The images \emph{baby} from Set5 \cite{set5} generated by DSN with or without Q-BReLU.}
\label{fig:showqbrelu}
\end{figure}

However, the gradient of Q-BReLU alternates between 0 and $\infty$ according to \eqref{qbrelu}. We exploit an approximate gradient with local continuity for backpropagation learning. To retain center quantization and zero mean deviation, BReLU is adopted as a spline function to fit Q-BReLU, as shown in Fig. \ref{subfig:gqbrelu}.
Therefore, the approximate gradient of Q-BReLU is defined as
\begin{equation}\label{gqbrelu}
\frac{{\partial {f_{qbrelu}}\left( x \right)}}{{\partial x}} = \left\{ \begin{array}{l}
1,\;{{t}_{\min }} < x < {{t}_{\max }}\\
0,\;otherwise
\end{array} \right..
\end{equation}
To verify the impact of Q-BReLU, we illustrate an example of $\times2$ down-sampling in Fig. \ref{fig:showqbrelu}. The LR image generated without Q-BReLU contains irrational noise, while the result with Q-BReLU appears natural.

\subsection{Residual Up-sampling Subnetwork}
 The up-sampling subnetwork (Up-SNet) is a dense architecture that learns the sub-pixel residual, achieving a high performance and efficiency without cheap interpolation.

\subsubsection{Dense Pixel Representation}
Up-SNet densely connects the pixel representation \cite{densenet} for pixel-wise prediction. Each layer produces $k$ feature maps, so it follows that the $l$-th layer has $k\times l$ input feature maps. In the deep connection layers, a large number of feature maps increase the computational cost and model size. \cite{reception} demonstrated that a $1 \times 1$ bottleneck convolution improves computational efficiency and keeps the model compact. The performance improvement of dense pixel representation come from three aspects:
\begin{enumerate}[(1)]
  \item Multi-scale feature. For up-scaling, different components may be relevant to different neighbourhood scales in the LR image. Up-SNet is a type of multi-scale, performance-improving architecture: the receptive field get larger when the network stacks more layers. Given a fixed kernel of size $3\times3$, there are multi-scale streams corresponding to $\{3, 5, 7, 9\}$ receptive fields, respectively.
  \item Deeply-supervised learning. Image up-sampling is a low-level vision task, where the kernels in the shallow layers can be shared to recursively boost performance. However, recursions are hard to train due to exploding and/or vanishing gradients. Skip connections, similar to the deeply-supervised learning, overcome the vanishing gradient problem and enhance feature propagation.
  \item Multi-level dictionaries. Sparse-coding is a representative example-based up-sampling method, where sparse coefficients are passed into a dictionary to restore HR patches. Up-SNet can be viewed as a type of sparse coding: convolutional kernels of size $3\times3$ are equivalent to dictionaries, and the bottlenecks with nonlinear activation functions are equivalent to sparse coefficients. With dense connections, the neural unit learns multi-level dictionaries. 
\end{enumerate}

\subsubsection{Sub-pixel Residual Learning}

The HR image can be decomposed into low-frequency information (low-resolution image) and high-frequency information (residual image). In Up-SNet, the input image $\textbf{L}$ and output image $\textbf{S}$ share the same low-frequency information. Without any cheap interpolation, we adopt a sub-pixel residual learning to transmit the LR input to the HR result.

Depending on different sub-pixel location in HR space, the residual patterns containing $s^2$ channels are activated by a convolution of size $1\times1$. Sub-pixel shuffle \cite{espcn} is a periodic operator that rearranges the elements of an $H \times W \times s^2$ tensor to a tensor of shape $s\cdot H \times s\cdot W$.
In the mathematica formula, the sub-pixel residual image for up-sampling is written as $\textbf{R}_{\left\lfloor {x/s} \right\rfloor ,\left\lfloor {y/s} \right\rfloor ,s \cdot \left( {y\backslash s} \right) + \left( {x\backslash s} \right)}^u = {\textbf{S}_{x,y}} - {\textbf{L}_{\left\lfloor {x/s} \right\rfloor ,\left\lfloor {y/s} \right\rfloor }}$, where $\backslash$ denotes the remainder operator. To learn the sub-pixel residual image similarity to \eqref{L}, the restoration result is defined by
 \begin{equation}\label{S}
 {\textbf{S}_{x,y}} = {\mathcal{F}^u}{\left[ \textbf{L} \right]_{\left\lfloor {x/s} \right\rfloor ,\left\lfloor {y/s} \right\rfloor ,s \cdot \left( {y\backslash s} \right) + \left( {x\backslash s} \right)}} + {\textbf{L}_{\left\lfloor {x/s} \right\rfloor ,\left\lfloor {y/s} \right\rfloor }},
 \end{equation}
where $\mathcal{F}^u{\left[ \cdot \right]}$ is the sub-pixel residual prediction. It is effectively implemented using a tile layer and an element-wise sum layer.

\section{Experiments}

\subsection{Implementation Details}
The model is trained on 91 images from \cite{sc} and 200 images from the training set in \cite{b100}, which are widely used for SR \cite{vdsr,drcn,drrn,memnet}. Following \cite{srcnn}, the luminance channel is only considered in YCbCr color space, because humans are more sensitive to luminance changes. We train a specific network for each scale factor ($\times 2, 3, 4$).

The detailed DSN configurations and parameter settings shown in Fig. \ref{fig:rrsr} are summarized in Table \ref{tab:rsrnet}. Motivated by the experiment, a leaky ReLU (LReLU) $f_{lrelu}(x) = \max\left({x, 0.05x}\right)$ is used instead of ReLU as the activation function, except in the output layer. The layers with residual learning are initialized by drawing randomly from a Gaussian distribution ($\mu = 0, \sigma = 0.001$), because most values in the residual images are likely to be zero or small. The other filter weights are initialized according to \cite{msra}.
\begin{table}[th]
\center
   \caption{The detailed configurations of DSN.}
    \label{tab:rsrnet}\includegraphics[width=0.65\linewidth]{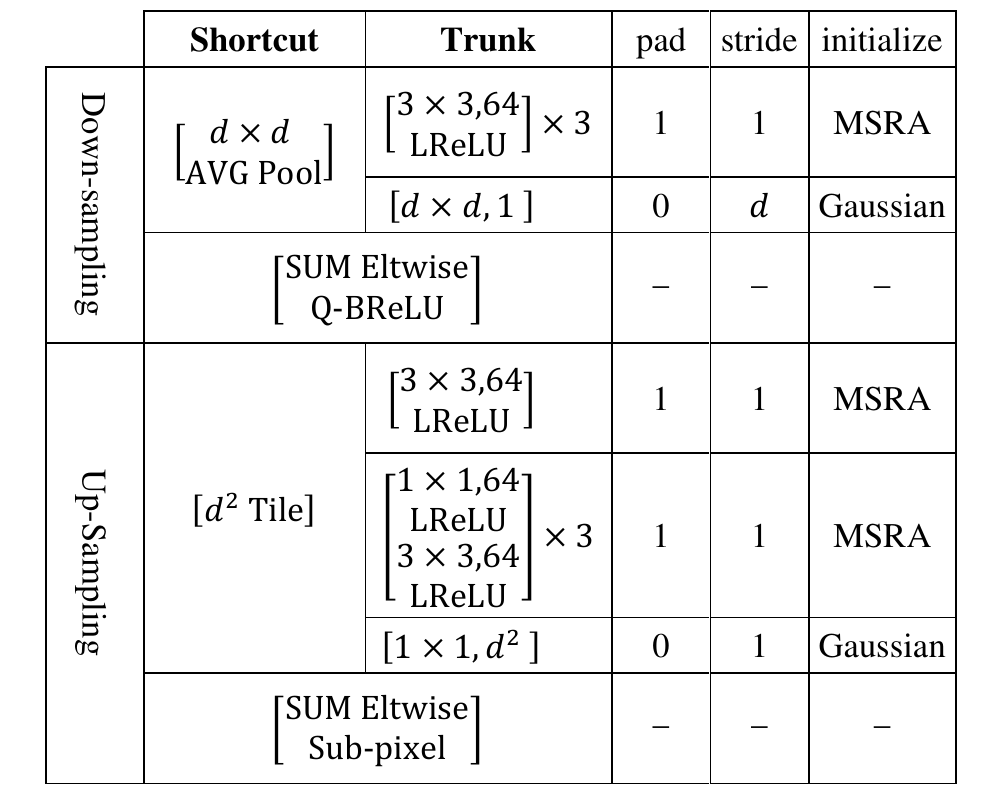}
\end{table}

 In the training phase, we rotate the images through $90^\circ$, $180^\circ$, and $270^\circ$ for data augmentation. Sub-images are extracted to ensure that all pixels in the original image appear once and only once as the ground truth of the training data. For $\times 2$, $\times3$, and $\times4$, we set the size of training sub-images as 60, 69, and 72, respectively. The model is trained with L1 loss using an Adam \cite{adam} optimizer in the \emph{Caffe} \cite{caffe} package. The learning rate decreases by half from $10^{-3}$ to $10^{-5}$ every 50 epochs. The final layer learns 10 times slower as in \cite{srcnn}. Based on the parameters above, training DSN with a batch-size of 256 takes about one day using one Nvidia GeForce GTX 1080 GPU.

\subsection{Image Reduced-Resolution Comparisons}

\begin{figure}[H]
\begin{center}
\includegraphics[width=0.5\linewidth]{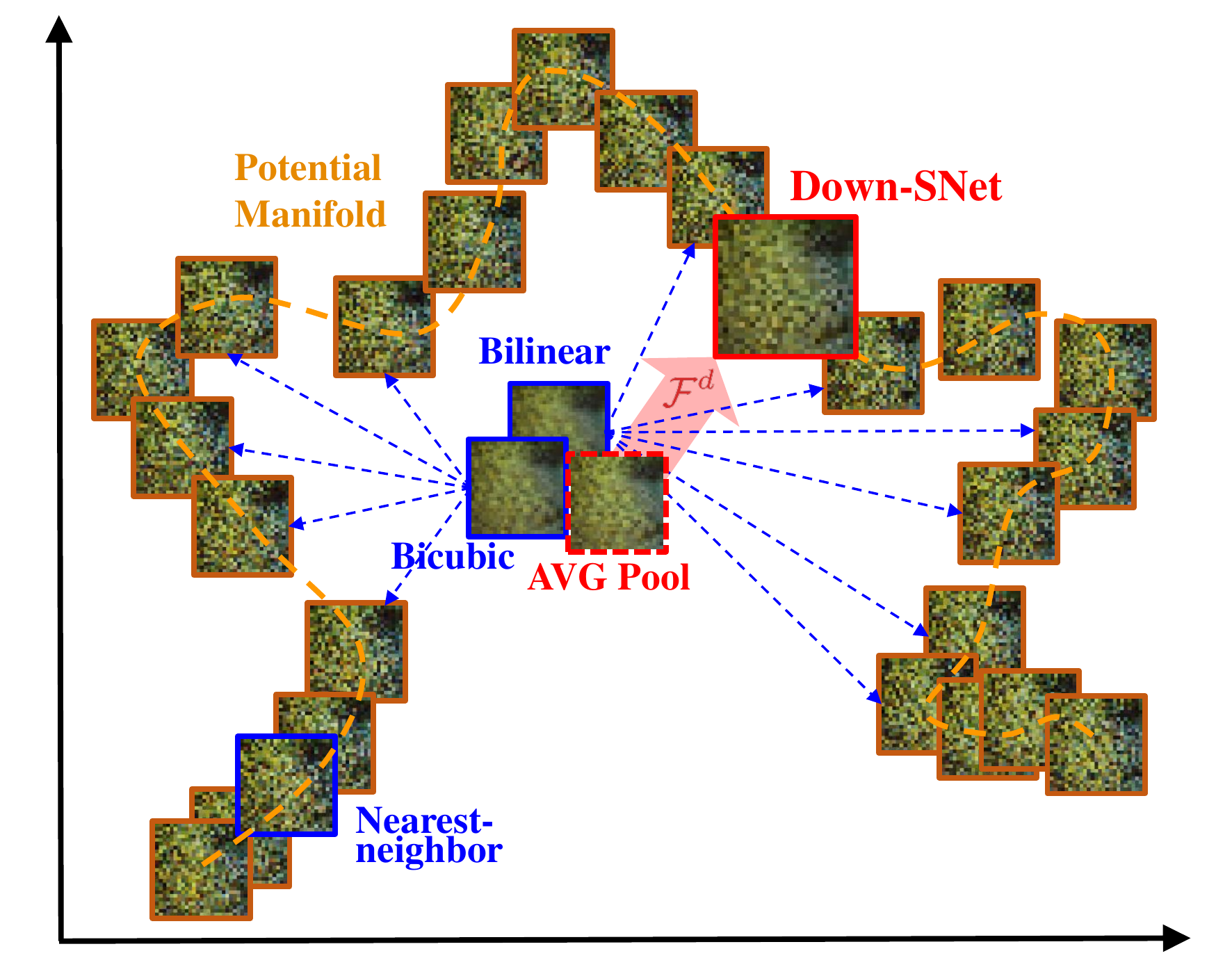}
\end{center}
   \caption{Illustration of potential solutions and LR results obtained with existing interpolations and Down-SNet.}
\label{fig:mainfold}
\end{figure}

Existing image down-sampling (e.g., nearest-neighbor, bilinear, and bicubic) is based on local weighting. Interpolation transformation struggles to find the pixel-wise weights of plausible solutions, which are typically over-smooth and of poor perceptual quality; that is, they will lose valuable high-frequency details such as texture. We illustrate this problem in Fig. \ref{fig:mainfold}, where multiple potential solutions with high textural details are weighted to create smooth bilinear or bicubic results. The solutions based on linear functions (bilinear and bicubic) appear overly smooth due to the pixel-wise weighing of possible solutions; the solution based on nearest-neighbor optionally selects a sample in the manifold space. While Down-SNet learns the residual $\mathcal{F}^d$ from a pixel-wise average (average pooling) towards the potential manifold, and produces perceptually more convincing solutions.

\begin{figure}[h]
\begin{center}
\includegraphics[width=0.18\linewidth]{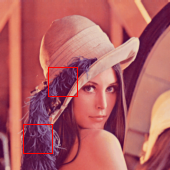}
\includegraphics[width=0.086\linewidth]{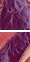}
\includegraphics[width=0.18\linewidth]{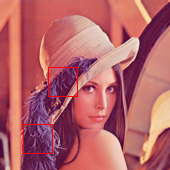}
\includegraphics[width=0.086\linewidth]{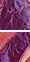}
\includegraphics[width=0.18\linewidth]{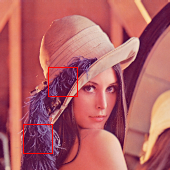}
\includegraphics[width=0.086\linewidth]{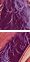}

\includegraphics[width=0.18\linewidth]{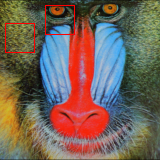}
\includegraphics[width=0.086\linewidth]{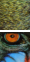}
\includegraphics[width=0.18\linewidth]{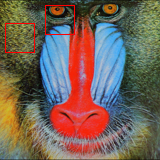}
\includegraphics[width=0.086\linewidth]{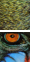}
\includegraphics[width=0.18\linewidth]{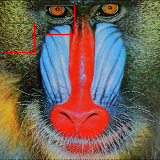}
\includegraphics[width=0.086\linewidth]{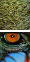}
\begin{tabular}{cC{3.3cm}C{3.3cm}C{3.3cm}c}
&(a) Bilinear&(b) Bicubic&(c) Down-SNet&
\end{tabular}
\end{center}
   \caption{Reduced-resolution results with scale factor $\times 3$. Down-SNet's result looks slightly sharper than bilinear and bicubic interpolation. (1) The first row shows the image \emph{lena}. Down-SNet generates more realistic and sharper hair textures. (2) The second row shows the image \emph{baboon}. Down-SNet produces high-frequency patterns missing in the bicubic and bilinear results, e.g., the fur and the light-spot in the baboon's eyeball.}
\label{fig:rrnet}
\end{figure}

The proposed method provides a powerful Down-SNet for generating photo-realistic LR images of high perceptual quality. Down-SNet encourages the LR image to move towards regions of the potential manifold with high probability of containing photo-realistic textures. Fig. \ref{fig:rrnet} shows two standard test images (\emph{lena} and \emph{baboon}) generated by Down-SNet compared to traditional down-sampling methods. Down-SNet generates relatively sharper and richer textures.

To quantitatively assess the down-sampling performance, bicubic degradation is replaced by Down-SNet to generate training samples for HR restoration. We retrain the SR networks with three types of representative architectures, including plain network (SRCNN \cite{srcnn_pami}), residual network (VDSR \cite{vdsr}), and dense network (Up-SNet). The Down-SNet remains fixed during the optimization process. Due to useful information preserving, Down-SNet brings significant improvement for HR restoration shown in Table \ref{tab:downsnet}. Although the Down-SNet is trained with Up-SNet in DSN, it has excellent generalization with the other network architectures. Therefore, Down-SNet as a CNN-based down-sampling can be used to replace traditional interpolation in image processing.
\begin{table}[h]
\center
\caption{Compare average PSNR with different down-sampling degradations for $\times 3$ SR on datasets Set5 \cite{set5}, Set14 \cite{set14}, B100 \cite{b100} and Urban \cite{selfex}.}
\scriptsize
\renewcommand\arraystretch{1.3}
\begin{tabular}{|c|C{1.4cm}|C{1.4cm}|C{1.4cm}|C{1.4cm}|C{1.4cm}|C{1.4cm}||c|}
\hline
\multirow{2}{*}{Dataset}&\multicolumn{2}{c|}{SRCNN \cite{srcnn_pami}}&\multicolumn{2}{c|}{VDSR \cite{vdsr}}& \multicolumn{2}{c||}{Up-SNet}&~~~DSN~~~\\
\cline{2-7}
&Bicubic&Down-SNet&Bicubic&Down-SNet&Bicubic&Down-SNet&~~~(co-train)~~~\\\hline
Set5&32.75&\textbf{33.24}&33.66&\textbf{33.85}&33.67&\textbf{33.98}&\textbf{34.29}\\
Set14&29.30&\textbf{29.78}&29.77&\textbf{29.93}&29.80&\textbf{30.19}&\textbf{30.30}\\
B100&28.41&\textbf{28.67}&28.82&\textbf{29.11}&28.58&\textbf{28.92}&\textbf{28.99}\\
Urban&26.24&\textbf{26.90}&27.14&\textbf{28.02}&26.81&\textbf{27.66}&\textbf{28.03}\\
\hline
\end{tabular}
\label{tab:downsnet}
\end{table}

\subsection{Image Super-Resolution Comparisons}

\begin{figure}[h]
\center
\scriptsize
\begin{tabular}{C{1.925cm}C{1.925cm}C{1.925cm}C{1.925cm}C{1.925cm}C{1.925cm}c}
Ground True&Bicubic&SRCNN \cite{srcnn_pami}&FSRCNN \cite{fsrcnn}&VDSR \cite{vdsr}&DSN&
\end{tabular}
\includegraphics[width=0.16\linewidth]{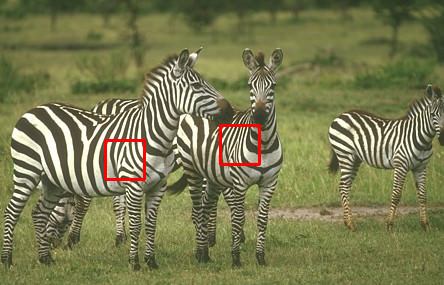}
\includegraphics[width=0.16\linewidth]{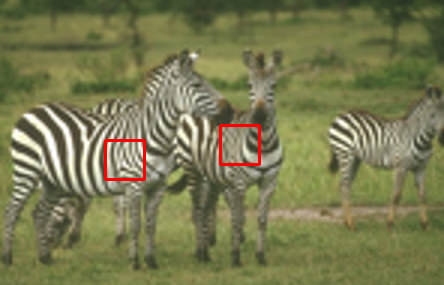}
\includegraphics[width=0.16\linewidth]{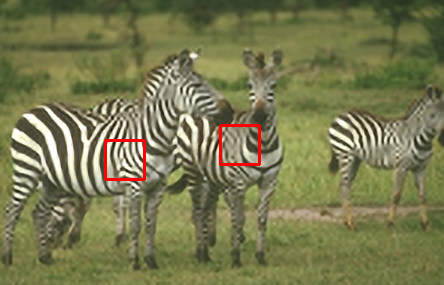}
\includegraphics[width=0.16\linewidth]{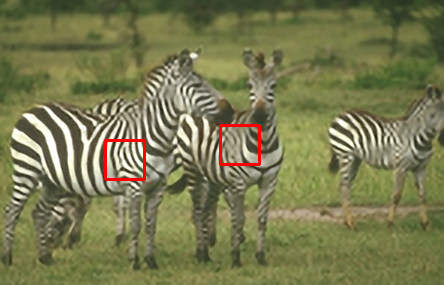}
\includegraphics[width=0.16\linewidth]{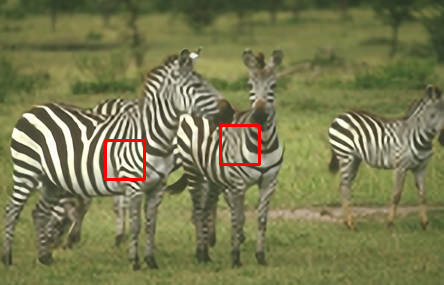}
\includegraphics[width=0.16\linewidth]{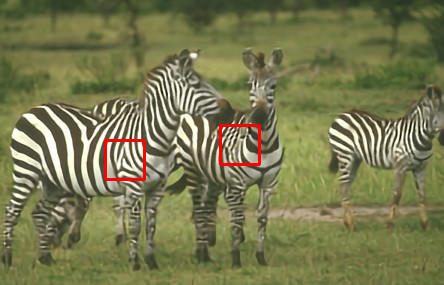}
\includegraphics[width=0.16\linewidth]{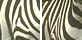}
\includegraphics[width=0.16\linewidth]{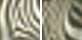}
\includegraphics[width=0.16\linewidth]{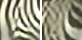}
\includegraphics[width=0.16\linewidth]{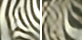}
\includegraphics[width=0.16\linewidth]{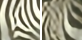}
\includegraphics[width=0.16\linewidth]{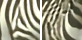}
\begin{tabular}{C{1.925cm}C{1.925cm}C{1.925cm}C{1.925cm}C{1.925cm}C{1.925cm}c}
(PSNR/SSIM)&(16.82/0.3572)&(19.93/\textcolor{blue}{\underline{0.4999}})&(19.62/0.4100)&(\textcolor{blue}{\underline{20.42}}/0.3765)&(\textcolor{red}{\textbf{22.44}}/\textcolor{red}{\textbf{0.7384}})&
\end{tabular}
\includegraphics[width=0.16\linewidth]{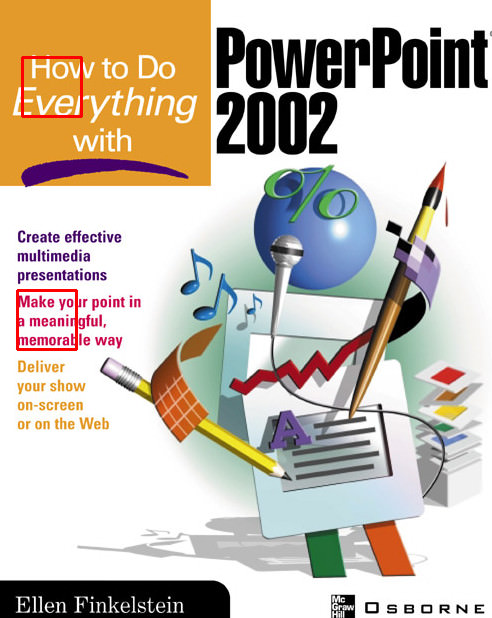}
\includegraphics[width=0.16\linewidth]{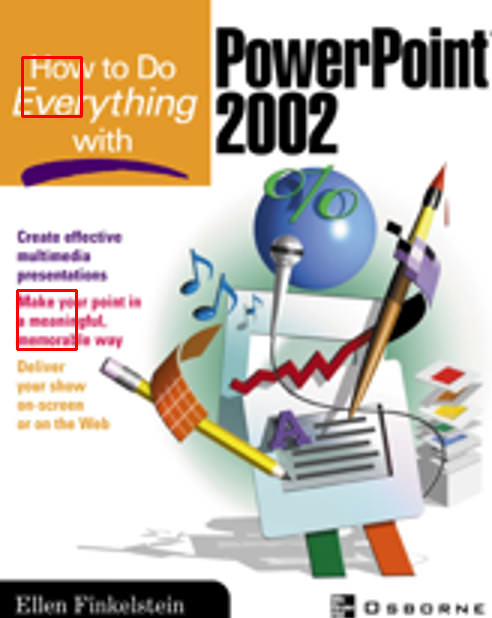}
\includegraphics[width=0.16\linewidth]{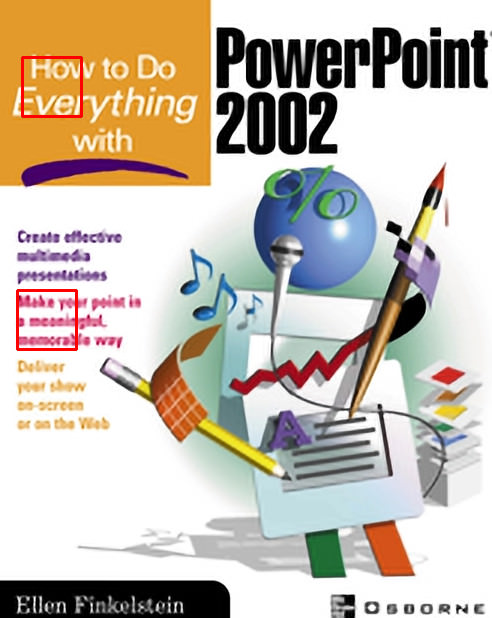}
\includegraphics[width=0.16\linewidth]{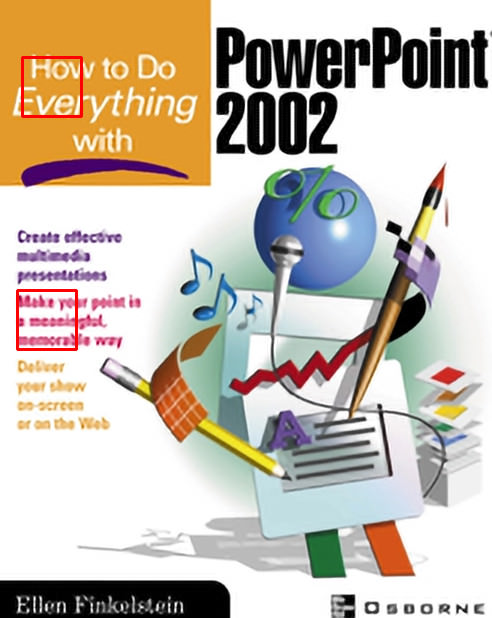}
\includegraphics[width=0.16\linewidth]{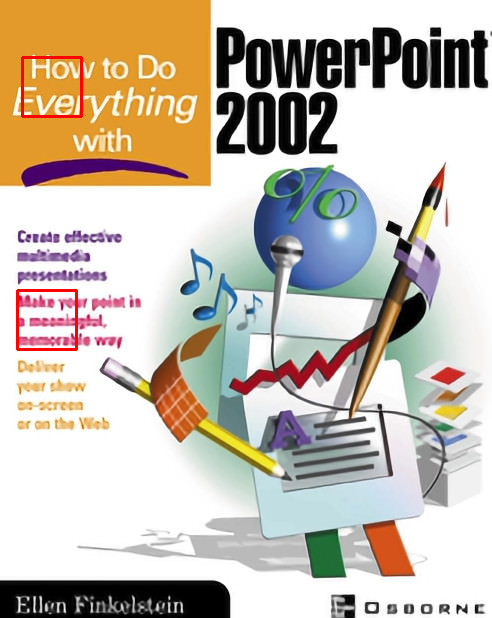}
\includegraphics[width=0.16\linewidth]{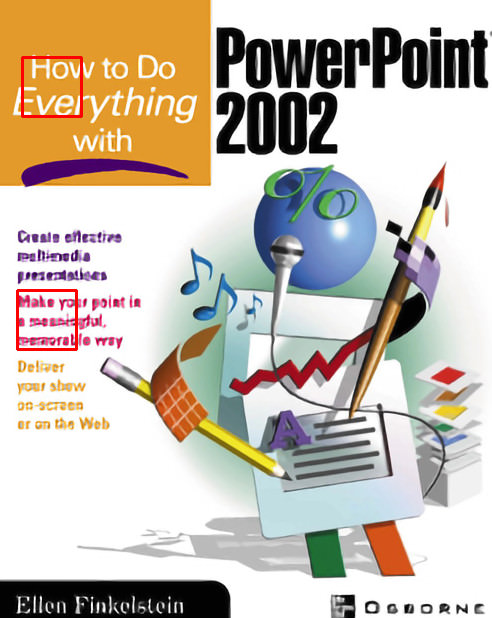}
\includegraphics[width=0.16\linewidth]{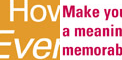}
\includegraphics[width=0.16\linewidth]{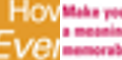}
\includegraphics[width=0.16\linewidth]{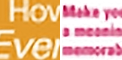}
\includegraphics[width=0.16\linewidth]{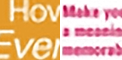}
\includegraphics[width=0.16\linewidth]{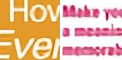}
\includegraphics[width=0.16\linewidth]{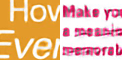}
\begin{tabular}{C{1.925cm}C{1.925cm}C{1.925cm}C{1.925cm}C{1.925cm}C{1.925cm}c}
(PSNR/SSIM)&(24.26/0.6997)&(26.19/0.7566)&(27.65/0.7637)&(\textcolor{blue}{\underline{30.27}}/\textcolor{blue}{\underline{0.7894}})&(\textcolor{red}{\textbf{32.37}}/\textcolor{red}{\textbf{0.8266}})&
\end{tabular}
\includegraphics[width=0.16\linewidth]{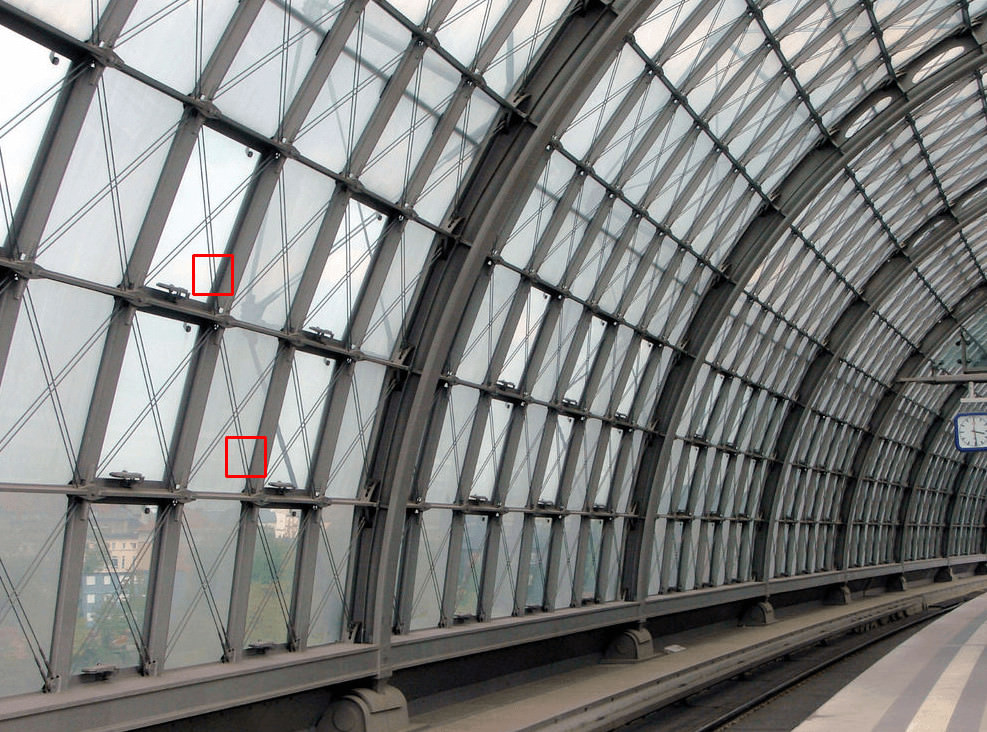}
\includegraphics[width=0.16\linewidth]{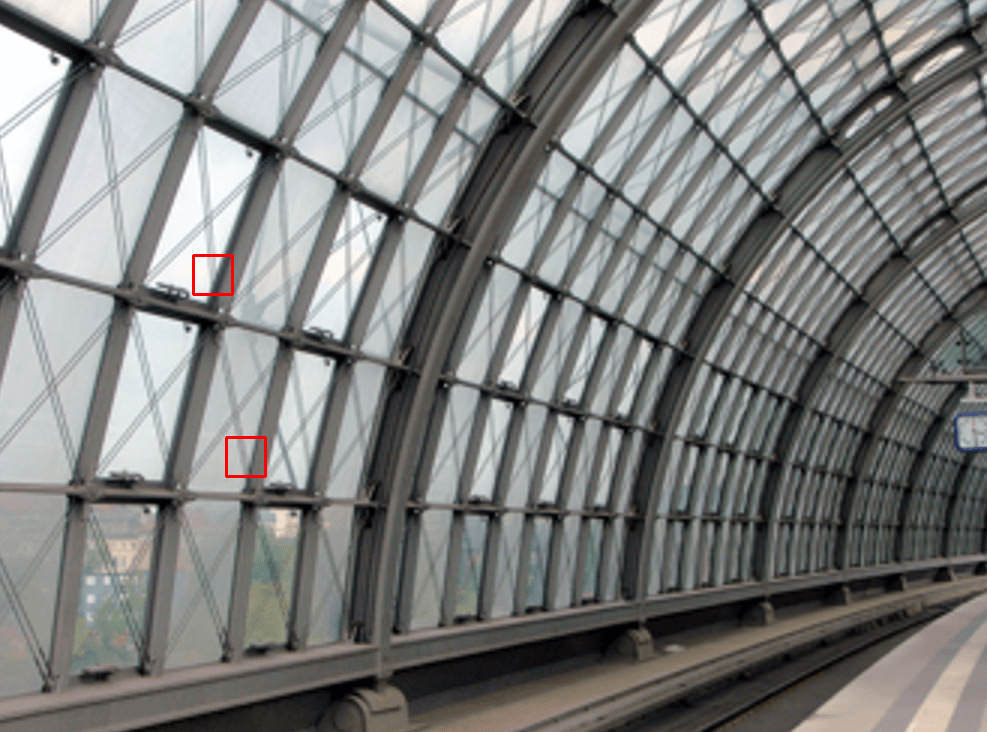}
\includegraphics[width=0.16\linewidth]{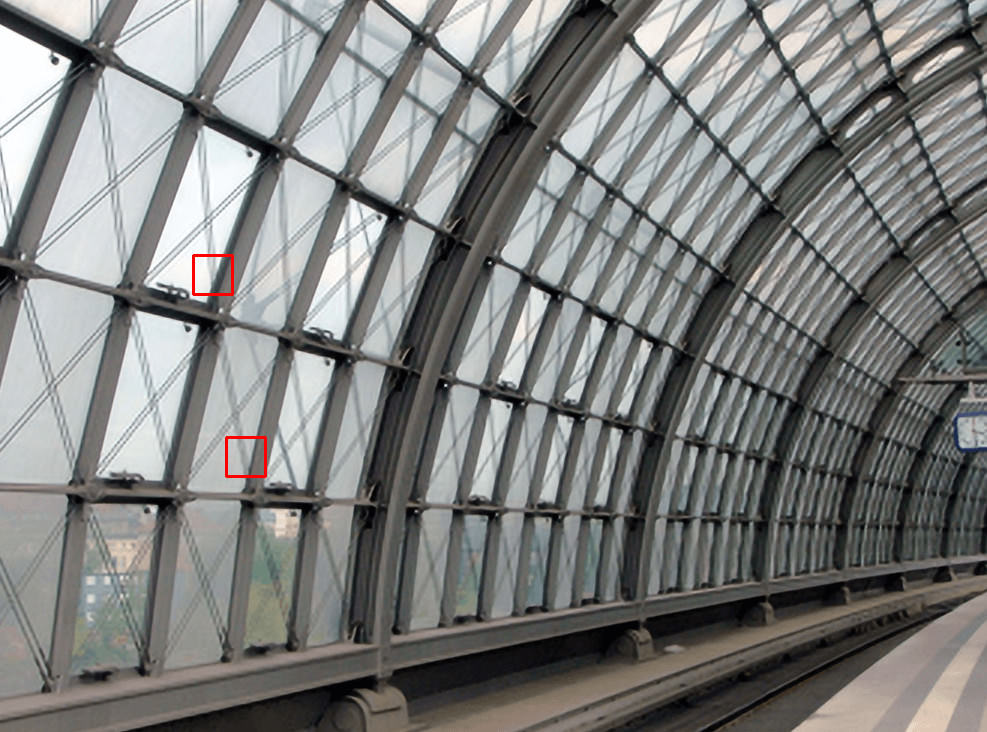}
\includegraphics[width=0.16\linewidth]{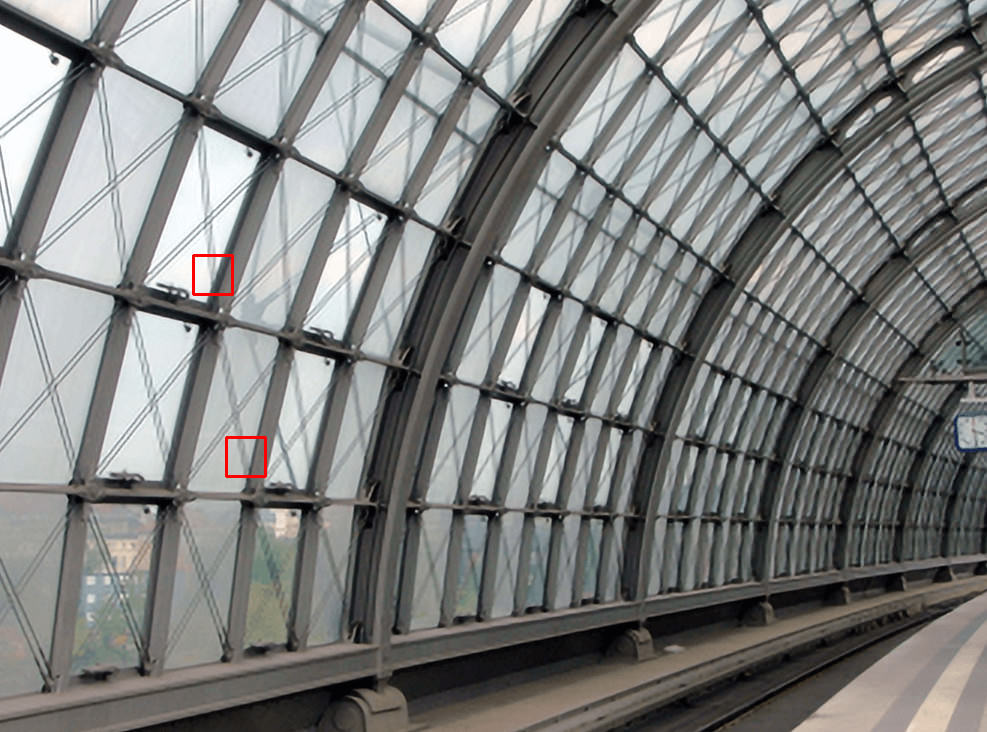}
\includegraphics[width=0.16\linewidth]{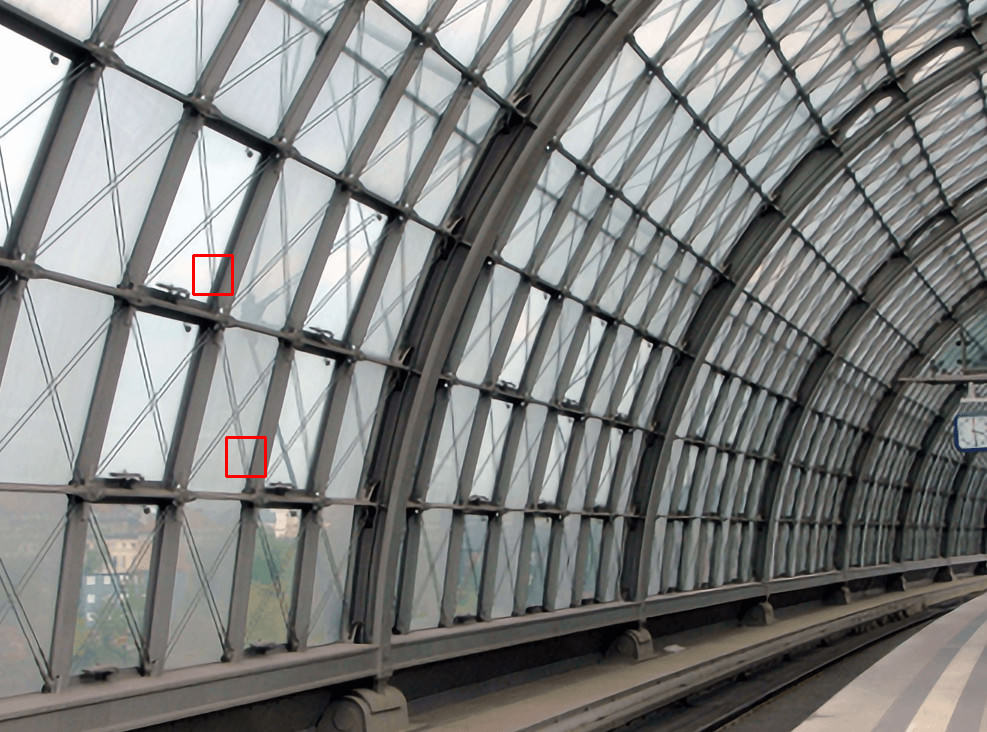}
\includegraphics[width=0.16\linewidth]{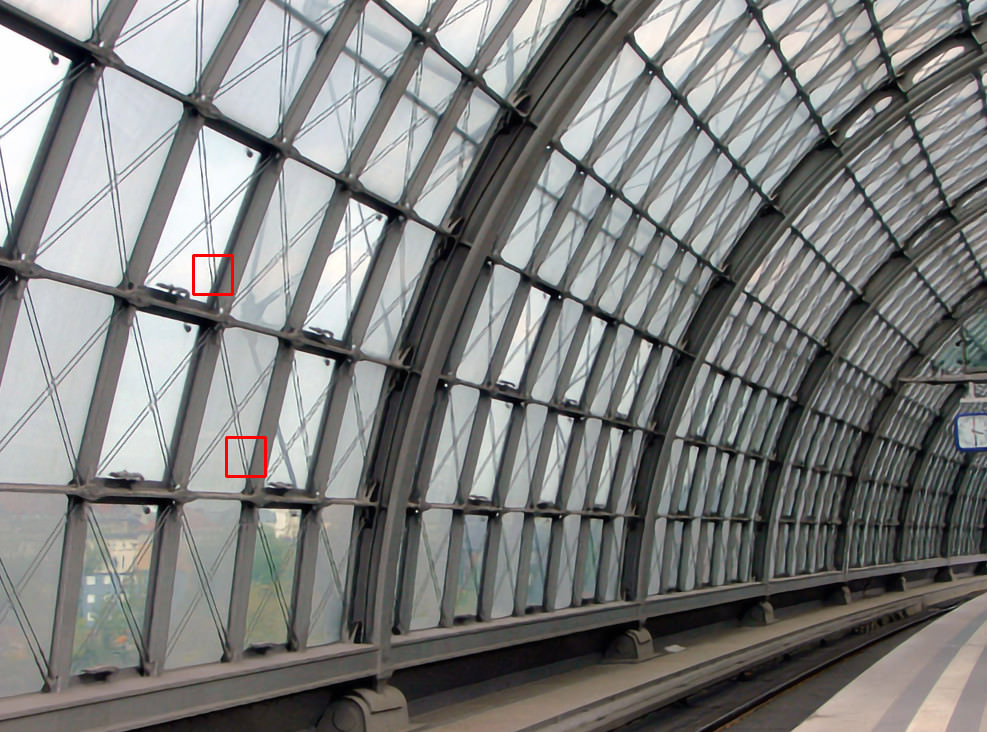}
\includegraphics[width=0.16\linewidth]{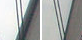}
\includegraphics[width=0.16\linewidth]{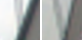}
\includegraphics[width=0.16\linewidth]{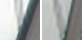}
\includegraphics[width=0.16\linewidth]{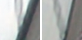}
\includegraphics[width=0.16\linewidth]{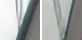}
\includegraphics[width=0.16\linewidth]{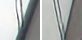}
\begin{tabular}{C{1.925cm}C{1.925cm}C{1.925cm}C{1.925cm}C{1.925cm}C{1.925cm}c}
(PSNR/SSIM)&(23.42/0.3567)&(23.95/0.0.3961)&(23.78/0.3567)&(\textcolor{blue}{\underline{25.24}}/\textcolor{blue}{\underline{0.4567}})&(\textcolor{red}{\textbf{27.92}}/\textcolor{red}{\textbf{0.9024}})&
\end{tabular}
   \caption{Super-resolution results with scale factor $\times 3$ and average PSNR/SSIM of each sub-figure. (1) The first row shows image \emph{253027} from B100 \cite{b100}. DSN accurately reconstruct the original pattern, while severe distortions are found in the results using other methods. (2) The second row shows image \emph{ppt3} from Set14 \cite{set14}. Text in DSN is sharp and identified, while others are blurry. (3) The last row shows image \emph{img002} from Urban \cite{selfex}. DSN reconstructs the lines well, while other methods generate blurry results.
    }
\label{fig:srimg}
\end{figure}

With the powerful down-sampling process, more useful information is preserved for image up-sampling. To further assess co-training DSN for SR, DSN is evaluated using three different scale factors ($\times 2$, $\times3$, $\times4$) on four datasets \cite{set5,set14,b100,selfex}. We compute the peak signal-to-noise ratio (PSNR) and structural similarity (SSIM) to compare five recent methods including FSRCNN \cite{fsrcnn}, VDSR \cite{vdsr}, DRCN \cite{drcn}, DRRN \cite{drrn}, and MemNet \cite{memnet}. As shown in Table \ref{tab:result}, the proposed DSN outperforms the other methods. Qualitative comparisons of SRCNN \cite{srcnn_pami}, FSRCNN \cite{fsrcnn}, and VDSR \cite{vdsr} are illustrated in Fig. \ref{fig:srimg} with their public codes. Our method produces relatively sharper edges and contours, while the other methods generate blurry results. In addition, existing methods produce severe distortions in some reconstructed results, whereas DSN reconstructs the texture patterns and avoids the distortions.
\begin{table}[bht]
\center
\caption{Average PSNR/SSIM for scale factors $\times 2$, $\times 3$ and $\times 4$ on datasets Set5 \cite{set5}, Set14 \cite{set14}, B100 \cite{b100} and Urban \cite{selfex}. \textcolor{red}{Red} indicates the best performance and \textcolor{blue}{blue} indicates the second-best performance.}
\scriptsize
\renewcommand\arraystretch{1.3}
\begin{tabular}{|c|c|c|c|c|c|c|c|}
\hline
Dataset&Scale&FSRCNN \cite{fsrcnn}&VDSR \cite{vdsr}&DRCN \cite{drcn}&DRRN \cite{drrn}&MemNet \cite{memnet}& DSN\\
\hline
\hline
\multirow{3}{*}{Set5}&$\times 2$&37.00/0.9558&37.53/0.9587&37.63/0.9588&37.74/\textcolor{blue}{\underline{0.9591}}&\textcolor{blue}{\underline{37.78}}/\textcolor{red}{\textbf{0.9597}}&\textcolor{red}{\textbf{37.92}}/0.9549\\
&$\times 3$&33.16/0.9140&33.66/0.9213&33.82/0.9226&34.03/0.9244&\textcolor{blue}{\underline{34.09}}/\textcolor{blue}{\underline{0.9248}}&\textcolor{red}{\textbf{34.29}}/\textcolor{red}{\textbf{0.9300}}\\
&$\times 4$&30.71/0.8657&31.35/0.8838&31.53/0.8854&31.68/0.8888&\textcolor{blue}{\underline{31.74}}/\textcolor{blue}{\underline{0.8893}}&\textcolor{red}{\textbf{31.92}}/\textcolor{red}{\textbf{0.9032}}\\
\hline
\hline
\multirow{3}{*}{Set14}&$\times 2$&32.63/0.9088&33.03/0.9124&33.04/0.9118&33.23/0.9136&\textcolor{blue}{\underline{33.28}}/\textcolor{blue}{\underline{0.9142}}&\textcolor{red}{\textbf{34.11}}/\textcolor{red}{\textbf{0.9286}}\\
&$\times 3$&29.43/0.8242&29.77/0.8314&29.76/0.8311&29.96/0.8349&\textcolor{blue}{\underline{30.00}}/\textcolor{blue}{\underline{0.8350}}&\textcolor{red}{\textbf{30.30}}/\textcolor{red}{\textbf{0.8578}}\\
&$\times 4$&27.59/0.7535&28.01/0.7674&28.02/0.7670&28.21/\textcolor{blue}{\underline{0.7721}}&\textcolor{blue}{\underline{28.26}}/\textcolor{red}{\textbf{0.7723}}&\textcolor{red}{\textbf{28.34}}/0.7539\\
\hline
\hline
\multirow{3}{*}{B100}&$\times 2$&31.50/0.8906&31.90/0.8960&31.85/0.8942&32.05/0.8973&\textcolor{blue}{\underline{32.08}}/\textcolor{blue}{\underline{0.8978}}&\textcolor{red}{\textbf{32.52}}/\textcolor{red}{\textbf{0.9074}}\\
&$\times 3$&28.52/0.7893&28.82/0.7976&28.80/0.7963&28.95/\textcolor{red}{\textbf{0.8004}}&\textcolor{blue}{\underline{28.96}}/\textcolor{blue}{\underline{0.8001}}&\textcolor{red}{\textbf{28.99}}/0.7969\\
&$\times 4$&26.96/0.7128&27.29/0.7251&27.23/0.7233&\textcolor{blue}{\underline{27.38}}/\textcolor{red}{\textbf{0.7284}}&\textcolor{red}{\textbf{27.40}}/\textcolor{blue}{\underline{0.7281}}&27.22/0.7010\\
\hline
\hline
\multirow{3}{*}{Urban}&$\times 2$&29.85/0.9009&30.76/0.9140&30.75/0.133&31.23/\textcolor{blue}{\underline{0.9188}}&\textcolor{blue}{\underline{31.31}}/0.9195&\textcolor{red}{\textbf{32.27}}/\textcolor{red}{\textbf{0.9305}}\\
&$\times 3$&26.42/0.8064&27.14/0.8279&27.15/0.8276&27.53/\textcolor{red}{\textbf{0.8378}}&\textcolor{blue}{\underline{27.56}}/\textcolor{blue}{\underline{0.8376}}&\textcolor{red}{\textbf{28.03}}/0.8346\\ &$\times 4$&24.60/0.7258&25.18/0.7524&25.14/0.7510&25.44/\textcolor{red}{\textbf{0.7638}}&\textcolor{blue}{\underline{25.50}}/\textcolor{blue}{\underline{0.7630}}&\textcolor{red}{\textbf{25.66}}/0.7145\\
\hline
\end{tabular}
\label{tab:result}
\end{table}

\begin{figure}[bht]
\begin{center}
   \includegraphics[width=0.5\linewidth]{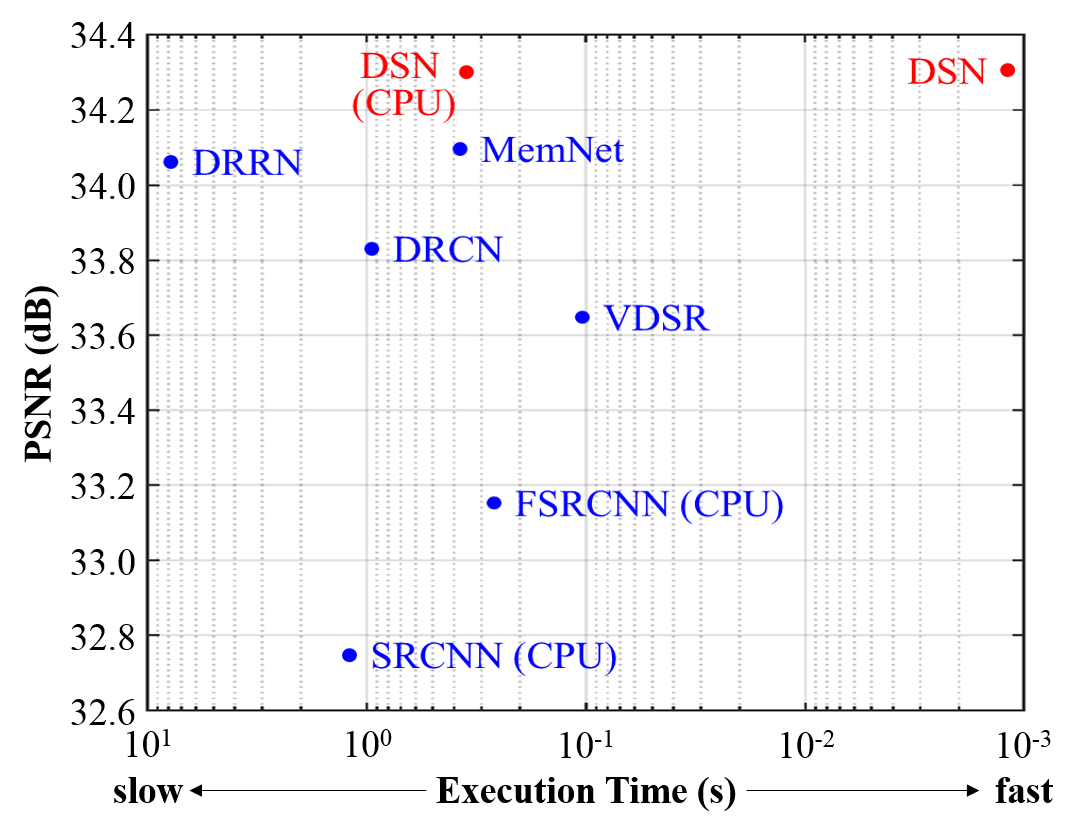}
\end{center}
   \caption{Plot of the trade-off between accuracy and speed for different methods on Set5 \cite{set5} with scale factor $\times 3$. The proposed DSN achieves better restoration quality than existing methods and is 100-times faster than MemNet \cite{memnet}. }
\label{fig:result}
\end{figure}

 In addition, we evaluate effectiveness according to execution time using the public code of the compared methods. The experiments are conducted with an Intel CPU (Xeon E5-2620, 2.1 GHz) and an NVIDIA GPU (GeForce GTX 1080). Fig. \ref{fig:result} shows the PSNR of the comparator methods versus execution time. The up-sampling phase of DSN out-performs existing methods. Even on a mobile CPU platform (A9 of iPhone 6S), our method for scale factor $\times3$ implemented with the \emph{ncnn}\footnote{https://github.com/Tencent/ncnn} library processes a $150 \times 150$ image in approximately 200 ms. Therefore, DSN can be used to generate thumbnails and reconstruct HR images for wide mobile applications.

\subsection{Image Compression Comparisons}

Image compression is a fundamental and well-studied engineering problem that aims to reduce irrelevance and redundancy for storage and transmission. With decreasing bits per pixel (bpp), high compression ratios cause blocking artifacts or noises in the decoded images. Existing image codecs usually consists of transformation, quantization, and entropy coding. Recently, deep learning-based image compression methods \cite{theis2017lossy,balle2016end} have achieved competitive performance. However, they are incompatible with existing image codecs, limiting their wideapread application in engineering. DSN is compatible with existing image coding standards to improve image compression. In DSN, Down-SNet produces a compact transformation for encoding using existing codecs, and Up-SNet reconstructs the decoded image to avoid blocking artifacts.

To evaluate the performance of DSN for image compression, we conduct experiments with standard compression methods including JPEG and JPEG2000. For compression evaluation, luminance values are usually considered in YCbCr color space. Bits for header information of compressed files count towards the bit rate of the compared methods. Since JPEG is without lossless compression, compared to JPEG, we use DSN for compression transforming and a common file compressor for quantization coding. The raw image is down-sampled by Down-SNet and stored as a \emph{.pgm} file, an uncompressed format. Then, the \emph{.pgm} file is coded by 7-Zip \footnote{http://www.7-zip.org/} with solid compression. Compared to JPEG2000, we simply adopt OpenJPEG \footnote{http://www.openjpeg.org/} with lossless compression to code the LR image generated by Down-SNet.
\begin{table}
\center
\caption{bpp/SSIM results of DSN + (7-Zip/JPEG2000), JPEG and JPEG2000 on dataset Set5 \cite{set5}.}
\scriptsize
\renewcommand\arraystretch{1.3}
\begin{tabular}{|c|C{2cm}|C{2cm}|C{2cm}|c|}
\hline
\multirow{2}{*}{Image}&\multicolumn{2}{c|}{JPEG}&\multicolumn{2}{c|}{JPEG2000 (J2K)}\\\cline{2-5}
&DSN + 7-Zip&JPEG ($q=44$)&DSN + J2K&~J2K ($r=13\%$)~~\\\hline
baby ($510\times510$)&0.6051/\textbf{0.9829}&\textbf{0.5367}/0.9614&\textbf{0.5189}/\textbf{0.9829}&0.6000/0.9713\\
bird ($288\times288$)&0.6726/\textbf{0.9656}&\textbf{0.6639}/0.9609&0.6012/\textbf{0.9656}&\textbf{0.5971}/0.9581\\
butterfly ($255\times255$)&\textbf{0.7420}/\textbf{0.9608}&1.1448/0.9411&0.7524/\textbf{0.9608}&\textbf{0.5878}/0.9160\\
head ($279\times279$)&\textbf{0.6050}/\textbf{0.8376}&0.6057/0.8082&\textbf{0.5400}/0.8376&0.5700/\textbf{0.8470}\\
woman ($228\times342$)&\textbf{0.6689}/\textbf{0.9439}&0.7252/0.9326&0.6157/\textbf{0.9439}&\textbf{0.5960}/0.9436\\\hline
Average&\textbf{0.6587}/\textbf{0.9300}&0.7353/0.9290&\textbf{0.5800}/\textbf{0.9300}&0.5902/0.9272\\\hline
\end{tabular}
\label{tab:compress}
\end{table}

\begin{figure}[th]
\center
\begin{tabular}{C{2.95cm}C{2.95cm}C{2.95cm}C{2.95cm}c}
Ground True&JPEG&JPEG2000&DSN&
\end{tabular}
\includegraphics[width=0.16\linewidth]{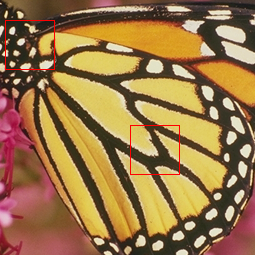}
\includegraphics[width=0.079\linewidth]{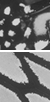}
\includegraphics[width=0.16\linewidth]{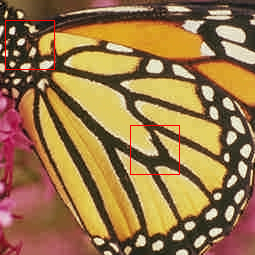}
\includegraphics[width=0.079\linewidth]{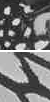}
\includegraphics[width=0.16\linewidth]{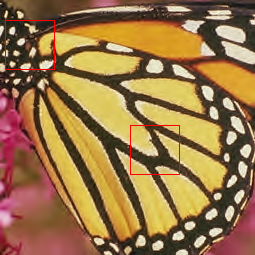}
\includegraphics[width=0.079\linewidth]{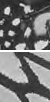}
\includegraphics[width=0.16\linewidth]{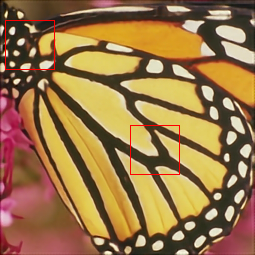}
\includegraphics[width=0.079\linewidth]{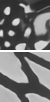}
\begin{tabular}{C{2.95cm}C{2.95cm}C{2.95cm}C{2.95cm}c}
(bbp/SSIM)&(0.7608/0.9450)&(0.7524/0.9353)&(\textbf{0.7420}/\textbf{0.9608})&
\end{tabular}
\caption{Subjective comparison of JPEG, JPEG2000, and DSN with bit rate 0.74bpp. Visually
disturbing blocking, aliasing, or ringing artifacts are commonly seen in images compressed with JPEG or JPEG2000. With DSN, compressed images preserves the smoothness and sharpness of many contours and edges, giving them a more natural appearance.}
\label{fig:compression}
\end{figure}

In Table \ref{tab:compress}, we evaluate the compression ratio on Set5 with a similar distortion factor SSIM. We use DSN trained with scale factor $\times 3$ as the transformation. For JPEG and JPEG2000, we test the codecs at quality parameter $q=44$ and compression ratio $r=13\%$, respectively. The comparisons show that the proposed method significantly outperforms JPEG and JPEG2000 in terms of bbp. To demonstrate the qualitative nature of compression artifacts, we show a representative example of the compressed image \emph{butterfly} with $\text{bpp}\approx 0.74$ in Fig. \ref{fig:compression}.

\section{Conclusion}
In image sampling, down-sampling loses useful information and up-sampling at the first layer does not provide extra information. To address these problems, hear we proposed a deep sampling network (DSN). DSN is an end-to-end system without any cheap interpolation to simultaneously learn mappings for resolution reduction and improvement. The down-sampling subnetwork in DSN can also be applied to generate photo-realistic LR images and replace traditional interpolation in image processing. Moreover, our experimental results reveal that the co-training network achieves state-of-the-art performance on SR at higher speed, and improves image compression with existing image coding standards.

\bibliographystyle{splncs}
\bibliography{egbib}
\end{document}